\documentclass[runningheads]{llncs}
\usepackage{amssymb}

\usepackage[FINAL,year=2026,ID=1605]{eccv}
\usepackage{eccv}

\usepackage{eccvabbrv}

\usepackage{graphicx}
\usepackage{booktabs}

\usepackage{amsmath}
\usepackage{amssymb}
\usepackage{amsfonts}
\usepackage{multirow}

\usepackage[accsupp]{axessibility}  

\usepackage{algorithm}
\usepackage{algorithmic} 

\usepackage{booktabs}   
\usepackage{multirow}  
\usepackage{pifont}     
\usepackage[table]{xcolor} 
\usepackage{amssymb}    
\usepackage{tabularx}
\usepackage{array}

\usepackage[utf8]{inputenc}

\usepackage{pgfplots}
\usepackage{xcolor}
\usepackage{colortbl}
\pgfplotsset{compat=1.18}

\definecolor{first}{RGB}{191, 225, 201} 
\definecolor{second}{RGB}{227, 237, 185} 
\definecolor{third}{RGB}{254, 250, 194} 
\definecolor{stab}{RGB}{232, 241, 252} 

\usepackage{hyperref}

\usepackage{orcidlink}

\usepackage{bbding}

\begin{document}

\title{This Looks Distinctly Like That: Grounding Interpretable Recognition in Stiefel Geometry against Neural Collapse} 

\titlerunning{This Looks Distinctly Like That}

\author{Junhao Jia\inst{1,3}\textsuperscript{\smash{$\dagger$}}\orcidlink{0009-0002-7508-4715} \and
Jiaqi Wang\inst{2}\textsuperscript{\smash{$\dagger$}}\orcidlink{0009-0000-1766-7774} \and
Yunyou Liu\inst{3}\orcidlink{0009-0006-0643-0601} \and
Haodong Jing\inst{4}\orcidlink{0000-0001-6643-7588} \and \\
Yueyi Wu\inst{3}\orcidlink{0009-0006-6543-9729} \and 
Xian Wu\inst{2}\textsuperscript{\smash{(\Envelope)}}\orcidlink{0000-0003-1118-9710} \and 
Yefeng Zheng\inst{1}\textsuperscript{\smash{(\Envelope)}}\orcidlink{0000-0003-2195-2847} 
}

\authorrunning{J.~Jia et al.}

\institute{Medical Artificial Intelligence Lab, Westlake University, Hangzhou, China \and
Tencent Jarvis Lab, Shenzhen, China \and
Hangzhou Dianzi University, Hangzhou, China \and
Xi'an Jiaotong University, Xi'an, China\\
\email{junhaojia530@gmail.com, zhengyefeng@westlake.edu.cn}
}

\maketitle

\begingroup
\renewcommand{\thefootnote}{}
\footnotetext{$^{\dagger}$~Equal contribution; \Envelope~Corresponding author.}
\endgroup

\begin{abstract}
Prototype networks provide an intrinsic case based explanation mechanism, but their interpretability is often undermined by prototype collapse, where multiple prototypes degenerate to highly redundant evidence. We attribute this failure mode to the terminal dynamics of Neural Collapse, where cross entropy optimization suppresses intra class variance and drives class conditional features toward a low dimensional limit. To mitigate this, we propose Adaptive Manifold Prototypes (AMP), a framework that leverages Riemannian optimization on the Stiefel manifold to represent class prototypes as orthonormal bases and make rank one prototype collapse infeasible by construction. AMP further learns class specific effective rank via a proximal gradient update on a nonnegative capacity vector, and introduces spatial regularizers that reduce rotational ambiguity and encourage localized, non overlapping part evidence. Extensive experiments on fine-grained benchmarks demonstrate that AMP achieves state-of-the-art classification accuracy while significantly improving causal faithfulness over prior interpretable models.


\end{abstract}

\section{Introduction}

\begin{figure}
    \centering
    \includegraphics[width=1\linewidth]{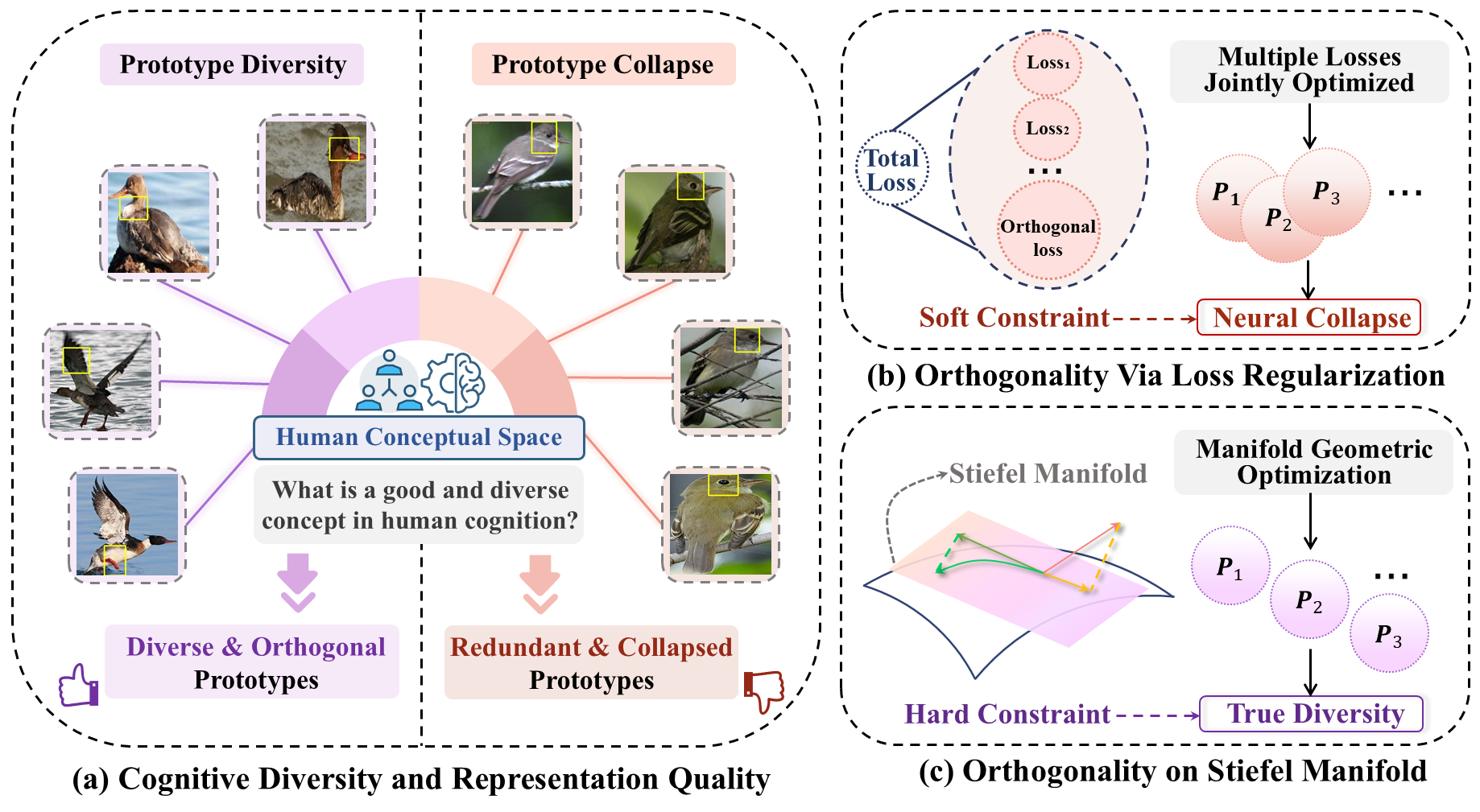}
    \caption{Motivation of Adaptive Manifold Prototypes (AMP). (a) Contrast between diverse and redundant prototypes. (b) Loss regularizations yield soft constraints. (c) AMP imposes a geometric constraint on the Stiefel manifold, ensuring part diversity.}
    \label{fig1}
\end{figure}

Deep visual recognition systems are increasingly deployed in high-stakes settings where authoritative decisions demand intrinsic transparency and verifiable reasoning~\cite{rudin2019stop}. Within expert domains, this cognitive process hinges on structural deconstruction: practitioners verify a concept by explicitly isolating distinct anatomical parts and evaluating their resemblance to established archetypal evidence~\cite{biederman1987recognition}. To emulate this human-like reasoning, prototype networks~\cite{chen2019looks} operationalize this intuitive framework by learning a diverse set of representative visual exemplars, which are subsequently matched against local spatial features extracted from the input image. Theoretically, this architecture builds a robust semantic bridge between abstract convolutional representations and human-understandable spatial evidence, decisively transforming opaque logit aggregations into a visually verifiable similarity-matching paradigm.

Despite their conceptual promise, existing prototype architectures systematically suffer from severe functional degradation in fine-grained visual recognition tasks, where categorical distinctions rely on subtle morphological cues~\cite{gautam2022protovae}. As shwon in Fig.~\ref{fig1}(a), empirical observations reveal a pervasive homogenization phenomenon known as prototype collapse~\cite{parise2025looks}. Ideally, a model should exhibit representational diversity, explicitly capturing a comprehensive array of distinct anatomical components~\cite{zhang2014part}. However, these freely optimized vectors frequently lose their intended structural diversity, collapsing to focus on the exact same highly discriminative spatial region. This yields redundant prototypes that entirely ignore the broader physical context. Ultimately, this localized overfitting structurally undermines the foundational premise of compositional interpretability, creating a gap between the model's representations and human cognition~\cite{hoffmann2021looks}.

We argue that pervasive prototype homogenization is not merely an architectural flaw, but a geometric inevitability. It stems directly from the algorithmic friction between interpretable representation learning and standard cross-entropy optimization. Recent theoretical analyses of deep neural networks formalize this dynamic as Neural Collapse~\cite{papyan2020prevalence,yaras2022neural}. During the terminal phase of training, the optimization landscape actively penalizes intra-class variance to maximize the inter-class decision margin. Driven by this overwhelming discriminative pressure, spatial features of a given category are aggressively compressed, converging toward a singular, highly symmetric mean vector in the latent space~\cite{zhu2021geometric}. This exposes a profound structural paradox: while compositional reasoning relies on internal feature diversity to isolate distinct physical parts, the overarching classification objective relentlessly destroys this variance, collapsing the learned manifold into a zero-dimensional terminal state~\cite{branson2014bird,markou2024guiding}.

To mitigate prototype redundancy, prior works commonly add auxiliary loss terms that softly penalize similarity among Euclidean prototypes (Fig.~\ref{fig1}(b)). While such penalties can encourage diversity early in training, they do not provide a feasibility guarantee: under strong cross-entropy gradients and the terminal dynamics associated with Neural Collapse, Euclidean prototypes may still become nearly collinear and concentrate on the same discriminative region.

In this paper, we propose Adaptive Manifold Prototypes (AMP), which replaces soft orthogonality penalties with a hard geometric constraint, as depicted in Fig.~\ref{fig1}(c). Specifically, AMP parameterizes class prototypes as an orthonormal basis on the Stiefel manifold, making the rank-1 prototype configuration infeasible by construction. In addition, AMP introduces two spatial gauge-fixing objectives (spatial entropy minimization and overlap suppression) whose purpose is not to enforce orthogonality, but to select a semantically stable and localized basis within the rotation-invariant subspace optimized by the classifier. Overall, our contributions can be summarized as follows:
\begin{itemize}
    \item We theoretically bridge prototype collapse with the terminal dynamics of Neural Collapse, revealing how standard cross-entropy optimization geometrically drives unconstrained prototypes toward low-rank degeneracy.
    \item We propose Adaptive Manifold Prototypes (AMP), which formulates prototypes as orthonormal bases on the Stiefel manifold and employs dynamic rank calibration alongside spatial regularizers to structurally guarantee diverse, localized part discovery.
    \item Extensive experiments on fine-grained benchmarks demonstrate that AMP achieves state-of-the-art classification accuracy while setting new standards for causal faithfulness and stability among intrinsically interpretable models.
\end{itemize}

\section{Related Work}

\subsection{Post-hoc Interpretability Methods}
Post-hoc explanation techniques have long served as the standard approach for understanding deep neural networks, attempting to reverse-engineer pre-trained black-box models without architectural modifications. Among these, widely used gradient-based methods like Grad-CAM aim to highlight discriminative spatial regions~\cite{selvaraju2017grad}, yet extensive sanity checks reveal that their saliency maps are often weakly tied to learned parameters, behaving more like generic edge detectors than faithful representations of internal reasoning~\cite{adebayo2018sanity,kindermans2019reliability}. Beyond qualitative concerns, benchmark-style evaluations further suggest that popular attribution methods are often no better than simple baselines at identifying truly important features~\cite{hooker2019benchmark,yeh2019fidelity}. Moreover, explanations themselves can be fragile or even arbitrarily manipulated by imperceptible perturbations that preserve model outputs, undermining their trustworthiness~\cite{ghorbani2019interpretation,dombrowski2019explanations}. Furthermore, perturbation-based surrogates like LIME and SHAP~\cite{ribeiro2016should,lundberg2017unified} suffer from severe vulnerabilities to adversarial scaffolding and rationalization, enabling biased models to output deceptive explanations by post-hoc explainers~\cite{slack2020fooling,aivodji2019fairwashing}. Absent any direct regularization of the training dynamics, post-hoc methods cannot stop the network from learning shortcut rules, non-robust features, or entangled and redundant representations~\cite{geirhos2020shortcut,alvarez2018towards,ilyas2019adversarial,locatello2019challenging}. This structural limitation prompts a necessary shift toward inherently interpretable architectures and concept-based reasoning pipelines~\cite{rudin2019stop}.

\subsection{Prototype-based Interpretability Methods}

To address the inherent unfaithfulness of post-hoc techniques, prototype networks explicitly embed a transparent, case-based reasoning process into the architecture~\cite{alvarez2018towards}. Originating with ProtoPNet~\cite{chen2019looks}, these models classify images by matching local patches to learned prototypical vectors. Owing to its intuitive transparency, this case-based paradigm has seen extensive applications beyond standard visual recognition, extending into diverse domains such as medical diagnosis~\cite{wang2025cross,jia2026unsupervised}, natural language processing~\cite{hong2023protorynet,wei-zhu-2025-protolens}, and time-series analysis~\cite{ni2021interpreting,peng2026protots}. Unconstrained prototypes naturally gravitate toward the single most discriminative region, prompting numerous extensions to enforce structural diversity via hierarchical trees~\cite{nauta2021neural}, spatial matching~\cite{donnelly2022deformable}, and nearest-neighbor alignments~\cite{ukai2023this}. Concurrently, many works rely on heuristic loss regularizers, such as soft orthogonality penalties or spatial separation margins, to encourage diverse anatomical part discovery~\cite{wang2021interpretable,huang2023evaluation,wang2025mixture,ayoobi2025protoargnet}. However, lacking absolute geometric boundaries, these soft constraints are systematically overpowered by the primary cross-entropy objective~\cite{hoffmann2021looks,gautam2022protovae}, inevitably driving multiple prototypes to homogenize and redundantly attend to identical spatial regions~\cite{parise2025looks}. This severe prototype collapse fatally undermines compositional interpretability, demonstrating that robust part discovery requires a fundamental geometric restructuring of the representation space rather than mere auxiliary penalties.

\subsection{Optimization on Orthogonal Manifolds}

Optimization under orthogonality constraints has a rich history on Grassmann and Stiefel manifolds, progressing from early Newton and conjugate gradient algorithms \cite{edelman1998geometry} to modern unified frameworks utilizing retractions and vector transports \cite{absil2008optimization}. This geometric foundation has been further adapted for large-scale applications via constraint-preserving feasible methods \cite{wen2013feasible} and stochastic Riemannian algorithms for data-driven learning \cite{bonnabel2013stochastic,zhang2016riemannian}.

In deep learning, orthogonality frequently serves as a structural prior to stabilize training and enhance generalization by preserving gradient norms and mitigating optimization instabilities. Applications traditionally impose these constraints directly on the model's parameters. This ranges from orthogonal weight normalization \cite{huang2018orthogonal} and convolutional regularizers \cite{bansal2018can} to geometry-aware optimizers \cite{roy2018geometry} that maintain weight matrices on the Stiefel manifold, sometimes employing efficient approximations like the Cayley transform \cite{Li2020Efficient}. Similarly, in the latent space, representation learning leverages orthogonal and decorrelated objectives to promote disentanglement, rigorously encouraging individual latent dimensions to capture statistically independent generative factors while significantly reducing feature redundancy \cite{pandey2022disentangled,cogswell2016decov}.

Unlike prior works that apply orthogonality to network weights or latent factors, AMP leverages these geometric principles to enforce a strict Stiefel constraint on class-specific prototype bases to counteract Neural Collapse-induced prototype collapse, coupling this with adaptive rank pruning and spatial regularization to translate orthogonality into localized, compositional explanations.

\section{Preliminaries on Prototypical Networks}

Let $\mathcal{D} = \{ (x_i, y_i) \}_{i=1}^N \subset \mathcal{X} \times \mathcal{Y}$ be the training dataset, where $\mathcal{X}$ is the image space and $\mathcal{Y} = \{1, 2, \dots, C\}$ denotes the label space. A prototype network employs a parameterized convolutional backbone $f_\theta: \mathcal{X} \to \mathbb{R}^{D \times H \times W}$ to project an input $x_i$ into a spatial feature tensor $F^{(i)} = f_\theta(x_i)$. We denote the local feature vector at spatial location $(h, w) \in \mathcal{H} \times \mathcal{W}$ as $F_{h,w}^{(i)} \in \mathbb{R}^D$, where $\mathcal{H} = \{1, \dots, H\}$ and $\mathcal{W} = \{1, \dots, W\}$. The model maintains a learnable prototype set $\mathcal{P} = \{ P_1, \dots, P_C \}$, where $P_c = [p_{c,1}, \dots, p_{c,K}] \in \mathbb{R}^{D \times K}$ aggregates $K$ latent basis vectors assigned to class $c$.

During forward propagation, the network evaluates the spatial correspondence between a prototype $p_{c,k}$ and the input $x_i$ via global spatial max-pooling over a localized similarity metric $g: \mathbb{R}^D \times \mathbb{R}^D \to \mathbb{R}$:

\begin{equation}
S_{c, k}(x_i) = \max_{(h,w) \in \mathcal{H} \times \mathcal{W}} g\left(F_{h,w}^{(i)}, p_{c, k}\right)
\end{equation}

Crucially, to bridge the latent geometry with human-understandable visual concepts, the network relies on a strict prototype projection mechanism. The optimal coordinate $(h^*, w^*) = \arg\max_{(h,w)} g(F_{h,w}^{(i)}, p_{c, k})$ precisely localizes the semantic part in the input. Concurrently, to ensure intrinsic transparency, each continuous prototype vector is explicitly projected onto the nearest spatial feature patch from the training subset of class $c$:

\begin{equation}
p_{c,k} \leftarrow \arg\min_{\substack{j \in \{1,\dots,N\} \\ s.t. \; y_j=c}} \min_{(h,w) \in \mathcal{H} \times \mathcal{W}} \left\| p_{c,k} - F_{h,w}^{(j)} \right\|_2
\end{equation}

This projection grounds the abstract basis vectors into discrete, visually verifiable exemplars. These semantic activation scores form a representation vector $\mathbf{S}(x_i) \in \mathbb{R}^{CK}$, which is projected into the categorical probability simplex $\Delta^{C-1}$ via a linear fully connected layer to output the final prediction.

\begin{figure}
    \centering
    \includegraphics[width=\linewidth]{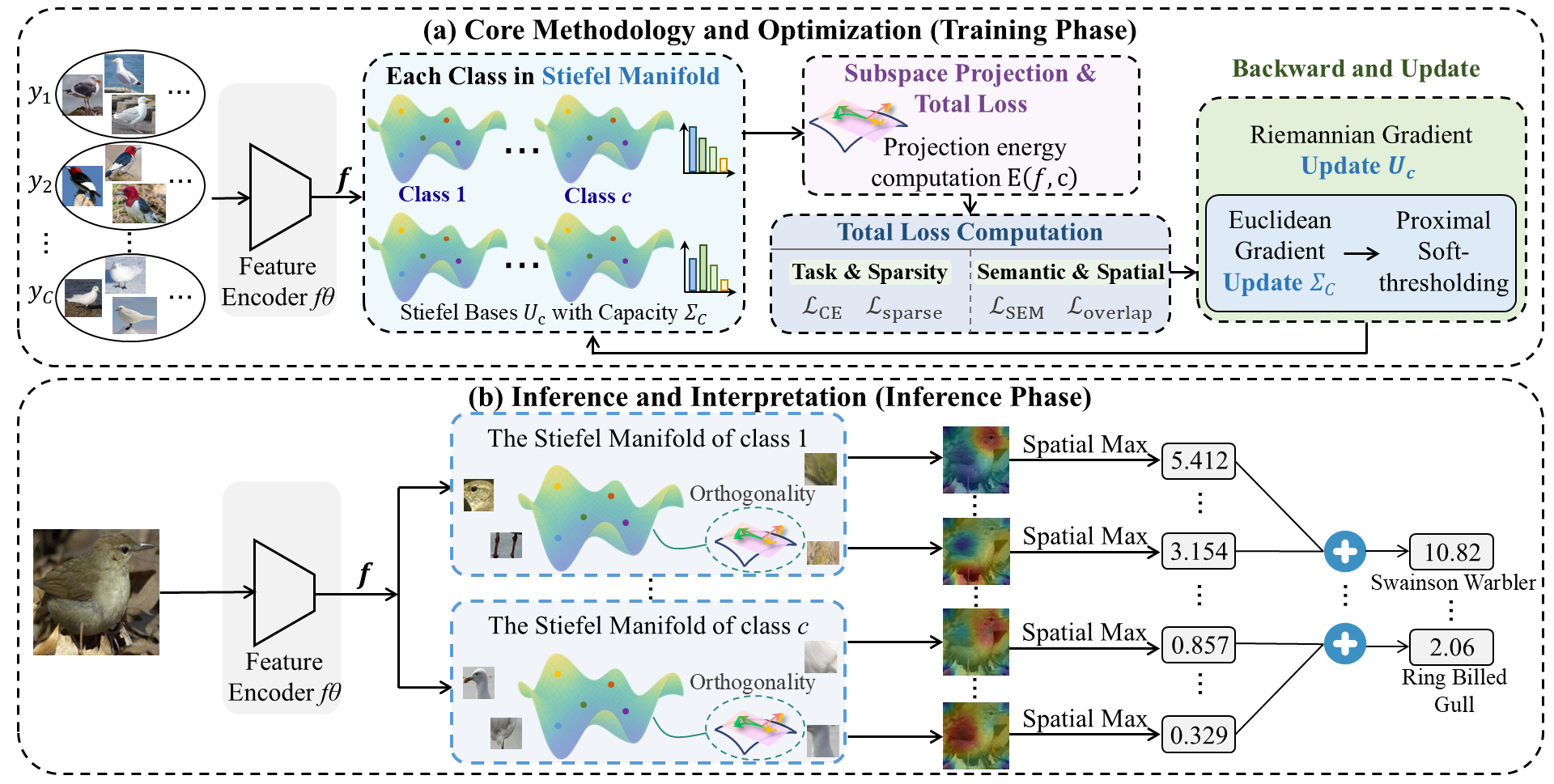}
    \caption{The overview of our proposed AMP framework. (a) Training: learn Stiefel bases and capacity matrices via Riemannian gradients. (b) Inference: project features onto class manifolds, then aggregate activations to class scores for interpretable predictions..}
    \label{fig2}
\end{figure}

\section{Methodology}

As illustrated in Fig.~\ref{fig2}, we propose the AMP framework, which replaces unconstrained Euclidean prototypes with Stiefel-constrained orthonormal bases for prototype matching, enabling diverse and collapse-free interpretable inference.

\subsection{Orthogonal Subspaces on the Stiefel Manifold}

Traditional prototype networks parameterize class visual features as unconstrained matrices $P_c \in \mathbb{R}^{D \times K}$. However, this formulation is structurally vulnerable to prototype collapse under the dynamics of Neural Collapse (NC)~\cite{papyan2020prevalence}. Let $\mu_c \in \mathbb{R}^D$ denote the global mean vector of class $c$, and $\Sigma_W$ represent the intra-class covariance matrix of the spatial features. During the terminal phase of cross-entropy optimization, NC dictates that $\lim_{t \to \infty} \Sigma_W^{(t)} = \mathbf{0}$, aggressively compressing all spatial features $\{F_{h,w}^{(i)} \mid y_i=c\}$ into the singular state $\mu_c$, while the class means self-organize into a Simplex Equiangular Tight Frame (ETF)~\cite{zhu2021geometric}.

To maximize the objective against this completely homogenized feature space, the unconstrained prototype matrix $P_c$ is forced to collapse, aligning all $K$ column vectors along the exact same mean coordinate:
\begin{equation}
\lim_{t \to \infty} p_{c, k}^{(t)} = \mu_c \quad \forall k \in \{1, \dots, K\}
\end{equation}

Geometrically, the effective rank of the class-specific prototype matrix fatally degenerates: $\lim_{t \to \infty} \text{rank}(P_c^{(t)}) = 1$. This zero-dimensional point collapse annihilates the local representational diversity required for compositional part discovery. To preclude this collapse, we abandon the unconstrained parameterization $P_c$ and strictly redefine the prototype set as an orthogonal basis matrix $U_c \in \mathbb{R}^{D \times K}$. We restrict $U_c$ to the Stiefel manifold $St(D, K)$~\cite{edelman1998geometry}:
\begin{equation}
St(D, K) = \{ U \in \mathbb{R}^{D \times K} \mid U^\top U = I_K \}
\end{equation}
By anchoring the prototypes to $St(D, K)$, we structurally endow the $K$ latent dimensions with absolute orthogonal separation. Strictly bound by the intrinsic curvature of the manifold, discriminative gradients are fundamentally prevented from compressing these basis vectors into a singular coordinate.

Under this rigid constraint, we replace the traditional Euclidean similarity metric with the projection energy onto the manifold's orthogonal subspace. Given a generic local feature vector $f \in \mathbb{R}^D$ (which represents any spatial feature $F_{h,w}^{(i)}$ from the input tensor), the mapping matrix projecting $f$ onto the subspace spanned by $U_c$ is $U_c U_c^\top$. The corresponding projection energy $E(f, c)$ is defined as the squared $\ell_2$ norm of this projection:
\begin{equation}
E(f, c) = \| U_c^\top f \|_2^2 = f^\top U_c U_c^\top f
\end{equation}

This paradigm shift fundamentally transforms the catastrophic point collapse (where rank degrades to $1$) into a controlled subspace representation (where rank is strictly preserved at $K$), geometrically guaranteeing the retention of representational capacity.

\subsection{Dynamic Rank Calibration via Proximal Gradients}

\label{Method2}

Real-world visual categories exhibit extreme asymmetry in their intrinsic semantic complexity. Imposing a uniform, fixed-rank subspace across all classes inevitably induces representational redundancy and overfitting. Motivated by the variable rank topology of nested subspaces~\cite{ye2022optimization}, we explicitly inject a learnable, non-negative diagonal capacity matrix $\Sigma_c = \text{diag}(\sigma_{c,1}, \dots, \sigma_{c,K})$ into the orthogonal basis. This operation transforms the rigid constraints of $St(D, K)$ into a dynamically weighted manifold framework defined by $F_c = U_c \Sigma_c$.

Within this dynamic scheme, the projection energy $E(f, c)$ mapping $f \in \mathbb{R}^D$ to the target class $c$ expands into an adaptively weighted quadratic form:
\begin{equation}
E(f, c) = f^\top U_c \Sigma_c U_c^\top f = \sum_{k=1}^K \sigma_{c,k} (U_{c,k}^\top f)^2
\end{equation}

To enforce the dimensional collapse of extraneous bases, we penalize the capacity matrix with a strict $\ell_1$ sparsity regularization. Given the non-negativity constraint on $\sigma_{c,k}$, this penalty simplifies to a direct linear summation:
\begin{equation}
\mathcal{L}_{sparse} = \lambda \sum_{c=1}^C \sum_{k=1}^K \sigma_{c,k}
\end{equation}

Standard stochastic gradient descent (SGD) is notoriously incapable of producing exact zero values for continuous parameters, leading to persistent, low-magnitude noise rather than true structural sparsity~\cite{shalev2011stochastic,langford2008sparse}. To enforce absolute dimensional collapse, we decouple the optimization and execute a proximal gradient descent step. Specifically, we apply a Euclidean soft-thresholding operator to the capacity weights at each iteration $t$:
\begin{equation}
\sigma_{c,k}^{(t+1)} = \max \left( \sigma_{c,k}^{(t)} - \eta \frac{\partial \mathcal{L}_{CE}}{\partial \sigma_{c,k}} - \eta \lambda, 0 \right)
\end{equation}

This exact truncation mechanism permits the effective subspace rank, denoted as $\text{rank}(F_c) = \| \text{diag}(\Sigma_c) \|_0$, to dynamically collapse to an optimal low-rank state. Such rigorous calibration ensures the active representational capacity perfectly mirrors the intrinsic physical complexity of each distinct visual category, completely suppressing the absorption of high-frequency noise.

\subsection{Semantic Gauge Fixing and Unsupervised Part Discovery}

Although the Stiefel constraint guarantees orthonormal columns for each class basis $U_c \in St(D,K)$, the classification score primarily depends on subspace geometry. Because an isotropic projector $U_c U_c^\top$ is invariant to right multiplication by any orthogonal matrix $Q \in O(K)$, and a learned diagonal capacity matrix $\Sigma_c$ only partially reduces this symmetry, rotational ambiguities persist. Consequently, the classification objective alone cannot guarantee a semantically stable basis. To mitigate this, we introduce two spatial regularizers that break rotational invariance, acting as an inductive bias to encourage individual basis directions to yield spatially localized and minimally overlapping visual evidence.

Given a spatial feature tensor $F \in \mathbb{R}^{D \times H \times W}$ for an input image, we compute a dense energy response map $M_{c,k,h,w} = \left(U_{c,k}^\top F_{h,w}\right)^2$ for each class $c$ and basis direction $k$, where $F_{h,w} \in \mathbb{R}^{D}$ is the local feature at spatial location $(h,w)$ and $U_{c,k}$ is the $k$th column of $U_c$. We then normalize this map into a spatial probability distribution via a spatial softmax:

\begin{equation}
P_{c,k,h,w} = \frac{\exp\!\left(M_{c,k,h,w}\right)}{\sum_{h^\prime=1}^{H}\sum_{w^\prime=1}^{W}\exp\!\left(M_{c,k,h^\prime,w^\prime}\right)}
\end{equation}

Rank calibration yields an active set of basis indices:

\begin{equation}
\mathcal{K}_c = \left\{ k \in \{1,\dots,K\} \mid \sigma_{c,k} > 0 \right\}, \qquad R_c = |\mathcal{K}_c|.
\label{eq:active_rank}
\end{equation}

The spatial regularizers are applied exclusively to this active set, which is treated as a discrete selection without backpropagating through its membership.

\paragraph{\textbf{Spatial Entropy Minimization.}}
To encourage focal and localized evidence for each active basis direction, we minimize the spatial entropy of $P_{c,k}$:
\begin{equation}
\mathcal{L}_{SEM}
=
-\frac{1}{R_c}
\sum_{k \in \mathcal{K}_c}
\sum_{h=1}^{H}\sum_{w=1}^{W}
P_{c,k,h,w}\log\!\left(P_{c,k,h,w}\right).
\label{eq:L_sem}
\end{equation}
Lower entropy corresponds to more concentrated maps, which empirically improves the localization quality of part evidence.

\paragraph{\textbf{Spatial Overlap Penalty.}}
Focal attention alone does not prevent multiple basis directions from attending to the same spatial region.
To encourage distinct evidence among active directions, we penalize pairwise overlap using cosine similarity between heatmaps:
\begin{equation}
\mathcal{L}_{overlap}
=
\frac{1}{R_c(R_c-1)}
\sum_{k \in \mathcal{K}_c}
\sum_{\substack{j \in \mathcal{K}_c \\ j \neq k}}
\frac{\langle P_{c,k}, P_{c,j} \rangle}{\|P_{c,k}\|_F \|P_{c,j}\|_F},
\label{eq:L_overlap}
\end{equation}
where $\langle \cdot, \cdot \rangle$ and $\|\cdot\|_F$ are the Frobenius inner product and Frobenius norm over spatial dimensions.
When $R_c < 2$, we set $\mathcal{L}_{overlap}$ to zero.

Together, these spatial objectives encourage the optimization to converge to a Stiefel basis whose columns correspond to focal and distinct spatial evidence.
This reduces rotational ambiguity in practice and improves semantic stability of part based explanations without requiring part annotations.

\subsection{Decoupled Optimization and Interpretable Inference}
\label{sec:optimization_inference}

We optimize AMP with a composite objective that combines classification, spatial regularization, and sparsity induced rank calibration:
\begin{equation}
\mathcal{L}_{total}
=
\mathcal{L}_{CE}
+
\gamma_1 \mathcal{L}_{SEM}
+
\gamma_2 \mathcal{L}_{overlap}
+
\lambda \sum_{c=1}^{C}\sum_{k=1}^{K}\sigma_{c,k},
\label{eq:L_total}
\end{equation}
subject to the constraints $U_c^\top U_c = I_K$ for all classes and $\sigma_{c,k} \ge 0$.
The parameters live on different spaces, so we use a decoupled update rule:
Euclidean updates for the backbone parameters $\theta$,
Riemannian updates for the Stiefel bases $U_c$,
and a proximal gradient update for the capacity weights $\sigma_{c,k}$.

AMP employs a decoupled optimization strategy to accommodate parameters residing in fundamentally different geometric spaces. While the feature backbone is updated using standard Euclidean optimizers, the class-specific orthogonal bases $U_c$ are optimized via Riemannian gradient descent, utilizing tangent space projections and QR-based retractions to strictly preserve the Stiefel manifold constraints at every iteration~\cite{absil2008optimization,edelman1998geometry}. Concurrently, the capacity matrix $\Sigma_c$ is updated using a Euclidean gradient step followed by a proximal soft-thresholding operation~\cite{parikh2014proximal}. This proximal step enforces exact $\ell_1$ sparsity on the continuous variables, physically driving the discrete dynamic rank calibration. 

\paragraph{\textbf{Interpretable inference.}}

Given a feature tensor $F^{(i)}$ for input $x_i$, we define the class logit as the maximum projection energy across spatial locations:
\begin{equation}
z_c(x_i) = \sum_{k=1}^K \sigma_{c,k} \left( \max_{(h,w)\in\mathcal{H}\times\mathcal{W}} (U_{c,k}^\top F_{h,w}^{(i)})^2 \right)
\label{eq:class_logit}
\end{equation}

The prediction is $\hat{c} = \arg\max_c z_c(x)$.
For explanation, we consider only the active directions $k \in \mathcal{K}_{\hat{c}}$, computing an energy heatmap:
\begin{equation}
\widetilde{M}_{\hat{c},k,i,j} = \sigma_{\hat{c},k}\left(U_{\hat{c},k}^\top F_{i,j}\right)^2,
\label{eq:inference_heatmap}
\end{equation}
and localize the most contributing region by $(h^\ast,w^\ast)=\arg\max_{i,j}\widetilde{M}_{\hat{c},k,i,j}$.
Finally, we retrieve a nearest training patch from the same class by searching over cached training features and selecting the patch that maximizes the corresponding energy response to ground each basis direction to a discrete exemplar:

\begin{equation}
F_{source} = \arg\max_{j \in \{1,\dots,N_{\hat{c}}\}, h, w} \left(U_{\hat{c},k}^\top F_{h,w}^{(j)}\right)^2
\end{equation}

\begin{table*}[t]
\centering
\caption{Predictive performance comparison (top-1 accuracy, \%) on CUB-200-2011 and Stanford Cars.
Best results among \textit{intrinsically interpretable models} are highlighted as \colorbox{first}{\textbf{first}}, \colorbox{second}{second} and \colorbox{third}{third}.
$^\dagger$ denotes ResNet50 pre-trained on iNaturalist for CUB.}
\label{tab:main_cub_cars_sanity_colored}

\setlength{\tabcolsep}{3.5pt}
\renewcommand{\arraystretch}{1.08}

\resizebox{\textwidth}{!}{%
\begin{tabular}{l cccc cccc}
\toprule
\multirow{2}{*}{\centering\raisebox{-0.65ex}{\textbf{Method}}}
& \multicolumn{4}{c}{\textbf{CUB-200-2011}}
& \multicolumn{4}{c}{\textbf{Stanford Cars}} \\
\cmidrule(lr){2-5}\cmidrule(lr){6-9}
& \textbf{V16} & \textbf{R34} & \textbf{R50$^\dagger$} & \textbf{D161}
& \textbf{V16} & \textbf{R34} & \textbf{R50} & \textbf{D161} \\
\midrule

Baseline
& 76.5 $\pm$ 0.8 & 80.0 $\pm$ 0.7 & 85.1 $\pm$ 0.5 & 84.5 $\pm$ 0.4
& 85.1 $\pm$ 0.4 & 86.7 $\pm$ 0.4 & 88.0 $\pm$ 0.3 & 89.0 $\pm$ 0.3 \\

\multicolumn{9}{c}{\cellcolor[HTML]{EEEEEE}\textit{Black-box Classifiers}} \\
RA-CNN
& 83.1 $\pm$ 0.6 & 84.9 $\pm$ 0.5 & 88.7 $\pm$ 0.4 & 88.0 $\pm$ 0.3
& 91.4 $\pm$ 0.3 & 92.2 $\pm$ 0.3 & 92.9 $\pm$ 0.2 & 93.4 $\pm$ 0.2 \\
NTS-Net
& 82.4 $\pm$ 0.6 & 84.1 $\pm$ 0.5 & 87.9 $\pm$ 0.4 & 87.2 $\pm$ 0.3
& 90.8 $\pm$ 0.3 & 91.7 $\pm$ 0.3 & 92.4 $\pm$ 0.2 & 92.9 $\pm$ 0.2 \\
PMG
& 84.7 $\pm$ 0.5 & 85.8 $\pm$ 0.5 & 89.2 $\pm$ 0.3 & 88.5 $\pm$ 0.3
& 92.0 $\pm$ 0.3 & 92.9 $\pm$ 0.2 & 93.6 $\pm$ 0.2 & 94.0 $\pm$ 0.2 \\

\multicolumn{9}{c}{\cellcolor[HTML]{EEEEEE}\textit{Intrinsically Interpretable Models}} \\
ProtoPNet
& 79.0 $\pm$ 0.7 & 80.5 $\pm$ 0.7 & 84.0 $\pm$ 0.5 & 83.3 $\pm$ 0.5
& 87.3 $\pm$ 0.4 & 88.0 $\pm$ 0.4 & 88.7 $\pm$ 0.3 & 89.5 $\pm$ 0.3 \\
TesNet
& 80.8 $\pm$ 0.6 & 82.1 $\pm$ 0.6 & 86.2 $\pm$ 0.4 & 85.3 $\pm$ 0.4
& 88.3 $\pm$ 0.4 & 89.2 $\pm$ 0.3 & 89.6 $\pm$ 0.3 & 90.0 $\pm$ 0.3 \\
ProtoKNN
& 80.3 $\pm$ 0.6 & 82.7 $\pm$ 0.6 & 84.9 $\pm$ 0.4 & 84.2 $\pm$ 0.4
& \cellcolor{third}88.4 $\pm$ 0.3 & \cellcolor{second}90.0 $\pm$ 0.3 & 89.9 $\pm$ 0.3 & \cellcolor{third}90.6 $\pm$ 0.2 \\
SDFA-SA
& 81.3 $\pm$ 0.6 & 83.0 $\pm$ 0.6 & \cellcolor{third}86.3 $\pm$ 0.6 & 85.1 $\pm$ 0.4
& 87.9 $\pm$ 0.4 & 88.7 $\pm$ 0.3 & 89.3 $\pm$ 0.3 & 89.9 $\pm$ 0.3 \\
MGProto
& \cellcolor{third}82.1 $\pm$ 0.6 & 83.6 $\pm$ 0.5 & \cellcolor{second}86.6 $\pm$ 0.4 & 85.2 $\pm$ 0.4
& \cellcolor{second}88.6 $\pm$ 0.3 & 89.7 $\pm$ 0.3 & \cellcolor{second}90.5 $\pm$ 0.2 & 90.5 $\pm$ 0.2 \\
CBC
& \cellcolor{second}82.4 $\pm$ 0.5 & \cellcolor{third}83.8 $\pm$ 0.5 & 86.0 $\pm$ 0.5 & \cellcolor{third}85.4 $\pm$ 0.4
& 88.2 $\pm$ 0.3 & \cellcolor{third}89.8 $\pm$ 0.3 & 90.0 $\pm$ 0.2 & \cellcolor{second}91.0 $\pm$ 0.2 \\
ProtoArgNet
& 82.0 $\pm$ 0.6 & \cellcolor{second}84.0 $\pm$ 0.5 & 85.8 $\pm$ 0.4 & \cellcolor{second}85.7 $\pm$ 0.4
& 88.3 $\pm$ 0.3 & 89.5 $\pm$ 0.3 & \cellcolor{third}90.3 $\pm$ 0.2 & 90.3 $\pm$ 0.2 \\
\midrule
\textbf{AMP (Ours)}
& \cellcolor{first}\textbf{84.2} $\pm$ 0.5 & \cellcolor{first}\textbf{85.5} $\pm$ 0.5 & \cellcolor{first}\textbf{88.4} $\pm$ 0.3 & \cellcolor{first}\textbf{87.8} $\pm$ 0.3
& \cellcolor{first}\textbf{90.4} $\pm$ 0.3 & \cellcolor{first}\textbf{91.3} $\pm$ 0.2 & \cellcolor{first}\textbf{92.0} $\pm$ 0.2 & \cellcolor{first}\textbf{92.5} $\pm$ 0.2 \\
\bottomrule
\end{tabular}%
}
\end{table*}

\section{Experiments}
\label{sec:experiments}

\subsection{Experimental Setup}

\paragraph{\textbf{Datasets and Metrics.}}
We conduct experiments on CUB-200-2011~\cite{wah2011caltech} and Stanford Cars~\cite{krause20133d}, two widely used benchmarks for fine-grained visual categorization. CUB-200-2011 contains 11,788 images from 200 bird species, while Stanford Cars contains 16,185 images from 196 car models. Following the established evaluation protocol~\cite{huang2023evaluation,wang2025mixture,jia2025geodesic}, we assess model performance by reporting Accuracy (Acc), Consistency (Cons.), Stability (Stab.), OIRR and DAUC.

\begin{figure}
    \centering
    \includegraphics[width=\linewidth]{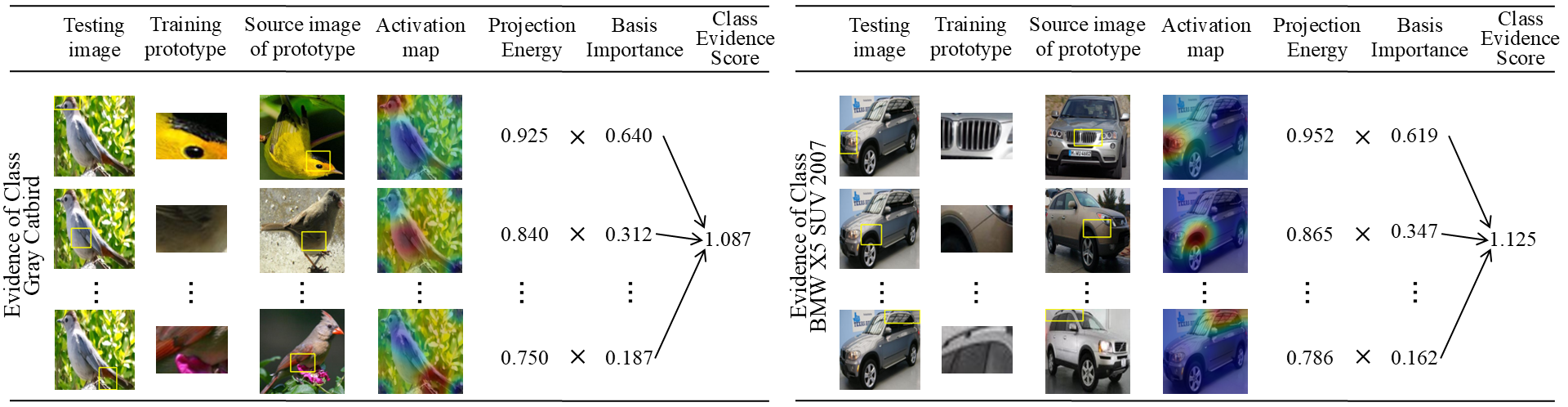}
    \caption{Qualitative visualization of the AMP reasoning process on the CUB-200-2011 (left) and Stanford Cars (right) datasets. For a given test image, AMP explicitly decomposes the final Class Evidence Score into a sparse sum of localized visual evidence.}
    \label{exp}
\end{figure}

\paragraph{\textbf{Baselines and Compared Methods.}}
As foundational convolutional backbones, we utilize VGG16, ResNet34, ResNet50, and DenseNet161 \cite{simonyan2015very,he2016deep,huang2017densely}, all initialized with ImageNet~\cite{deng2009imagenet} weights with the exception of the ResNet50 model on CUB, which employs iNaturalist~\cite{van2018inaturalist} pre-training. Furthermore, we benchmark our approach against representative black box fine grained classifiers (RA-CNN \cite{fu2017look}, NTS-Net \cite{yang2018learning}, and PMG \cite{du2020fine}) alongside several intrinsically interpretable models, including ProtoPNet \cite{chen2019looks}, TesNet \cite{wang2021interpretable}, ProtoKNN \cite{ukai2023this}, SDFA-SA \cite{huang2023evaluation}, MGProto \cite{wang2025mixture}, CBC \cite{saralajew2025robust}, and ProtoArgNet \cite{ayoobi2025protoargnet}.

\paragraph{\textbf{Implementation Details.}}
We implement AMP in PyTorch on a single NVIDIA RTX 4090 GPUs by resizing images to 448*448 pixels and cropping via bounding boxes. Backbones follow a decoupled optimization strategy where backbone parameters use SGD while Stiefel bases employ Riemannian SGD with QR retraction~\cite{absil2008optimization}. Initial prototype count $K$ is 10 for each class with sparsity $\lambda$ at 0.0001 and spatial weights $\gamma_1$ and $\gamma_2$ both at 0.01. The model undergoes training for 100 epochs with a batch size of 32 using a cosine annealing scheduler to decay the learning rate from 0.001 to 0.00001.

\begin{table*}[t]
\centering
\caption{Interpretability evaluation on CUB-200-2011 and Stanford Cars. Best results are highlighted as \colorbox{first}{\textbf{first}}, \colorbox{second}{second} and \colorbox{third}{third}.}
\label{tab:interpretability_metrics_sanity_colored}

\setlength{\tabcolsep}{3.5pt}
\renewcommand{\arraystretch}{1.08}

\resizebox{\textwidth}{!}{%
\begin{tabular}{l cccc cccc}
\toprule
\multirow{2}{*}{\centering\raisebox{-0.65ex}{\textbf{Method}}}
& \multicolumn{4}{c}{\textbf{CUB-200-2011}}
& \multicolumn{4}{c}{\textbf{Stanford Cars}} \\
\cmidrule(lr){2-5}\cmidrule(lr){6-9}
& \textbf{Cons.} & \textbf{Stab.} & \textbf{OIRR} & \textbf{DAUC}
& \textbf{Cons.} & \textbf{Stab.} & \textbf{OIRR} & \textbf{DAUC} \\
\midrule

\multicolumn{9}{c}{\cellcolor[HTML]{EEEEEE}\textit{Black-box Classifiers}} \\
RA-CNN
& 23.40$_{\pm 1.90}$ & 36.70$_{\pm 1.60}$ & 51.60$_{\pm 1.40}$ & 8.05$_{\pm 0.58}$
& 15.30$_{\pm 1.95}$ & 59.60$_{\pm 1.35}$ & 56.20$_{\pm 1.30}$ & 10.35$_{\pm 0.60}$ \\
NTS-Net
& 24.60$_{\pm 2.05}$ & 37.20$_{\pm 1.90}$ & 50.40$_{\pm 1.70}$ & 7.82$_{\pm 0.61}$
& 16.50$_{\pm 2.10}$ & 60.90$_{\pm 1.55}$ & 54.90$_{\pm 1.55}$ & 9.95$_{\pm 0.58}$ \\
PMG
& 26.00$_{\pm 2.25}$ & 38.60$_{\pm 2.15}$ & 47.90$_{\pm 2.00}$ & 7.35$_{\pm 0.60}$
& 18.10$_{\pm 2.30}$ & 62.30$_{\pm 1.90}$ & 53.20$_{\pm 1.95}$ & 9.38$_{\pm 0.56}$ \\

\multicolumn{9}{c}{\cellcolor[HTML]{EEEEEE}\textit{Intrinsically Interpretable Models}} \\
ProtoPNet
& 56.00$_{\pm 1.70}$ & 42.00$_{\pm 1.60}$ & 39.00$_{\pm 1.70}$ & 6.20$_{\pm 0.50}$
& 26.50$_{\pm 2.10}$ & 69.60$_{\pm 1.30}$ & 49.50$_{\pm 1.40}$ & 8.70$_{\pm 0.51}$ \\
TesNet
& 61.50$_{\pm 1.60}$ & 41.80$_{\pm 1.70}$ & 37.00$_{\pm 1.70}$ & 5.45$_{\pm 0.47}$
& 31.00$_{\pm 2.00}$ & 70.70$_{\pm 1.40}$ & 45.80$_{\pm 1.60}$ & 7.80$_{\pm 0.48}$ \\
ProtoKNN
& 63.00$_{\pm 1.80}$ & 42.60$_{\pm 1.90}$ & 35.20$_{\pm 1.90}$ & \cellcolor{second}3.80$_{\pm 0.45}$
& 34.50$_{\pm 2.20}$ & 71.40$_{\pm 1.60}$ & 42.60$_{\pm 1.80}$ & \cellcolor{third}6.15$_{\pm 0.46}$ \\
MGProto
& \cellcolor{second}71.40$_{\pm 1.90}$ & 45.80$_{\pm 2.30}$ & \cellcolor{third}30.80$_{\pm 2.10}$ & 4.02$_{\pm 0.44}$
& \cellcolor{second}45.00$_{\pm 2.40}$ & \cellcolor{third}74.20$_{\pm 2.00}$ & 36.80$_{\pm 2.10}$ & \cellcolor{second}6.05$_{\pm 0.43}$ \\
CBC
& \cellcolor{third}69.00$_{\pm 2.20}$ & \cellcolor{second}46.80$_{\pm 2.20}$ & 31.20$_{\pm 2.20}$ & 3.92$_{\pm 0.45}$
& \cellcolor{third}42.00$_{\pm 2.50}$ & \cellcolor{second}75.00$_{\pm 2.10}$ & \cellcolor{second}35.90$_{\pm 2.20}$ & 6.18$_{\pm 0.46}$ \\
ProtoArgNet
& 67.50$_{\pm 2.20}$ & \cellcolor{third}46.20$_{\pm 2.30}$ & \cellcolor{second}30.60$_{\pm 2.10}$ & \cellcolor{third}3.88$_{\pm 0.46}$
& 41.00$_{\pm 2.40}$ & 73.80$_{\pm 2.20}$ & \cellcolor{third}36.20$_{\pm 2.20}$ & 6.25$_{\pm 0.47}$ \\
\midrule
AMP (Ours)
& \cellcolor{first}\textbf{76.80}$_{\pm 1.80}$ & \cellcolor{first}\textbf{49.20}$_{\pm 2.10}$ & \cellcolor{first}\textbf{28.10}$_{\pm 2.00}$ & \cellcolor{first}\textbf{3.45}$_{\pm 0.41}$
& \cellcolor{first}\textbf{50.20}$_{\pm 2.20}$ & \cellcolor{first}\textbf{76.40}$_{\pm 1.90}$ & \cellcolor{first}\textbf{33.50}$_{\pm 2.00}$ & \cellcolor{first}\textbf{5.65}$_{\pm 0.40}$ \\
\bottomrule
\end{tabular}%
}
\end{table*}

\begin{figure}
    \centering
    \includegraphics[width=\linewidth]{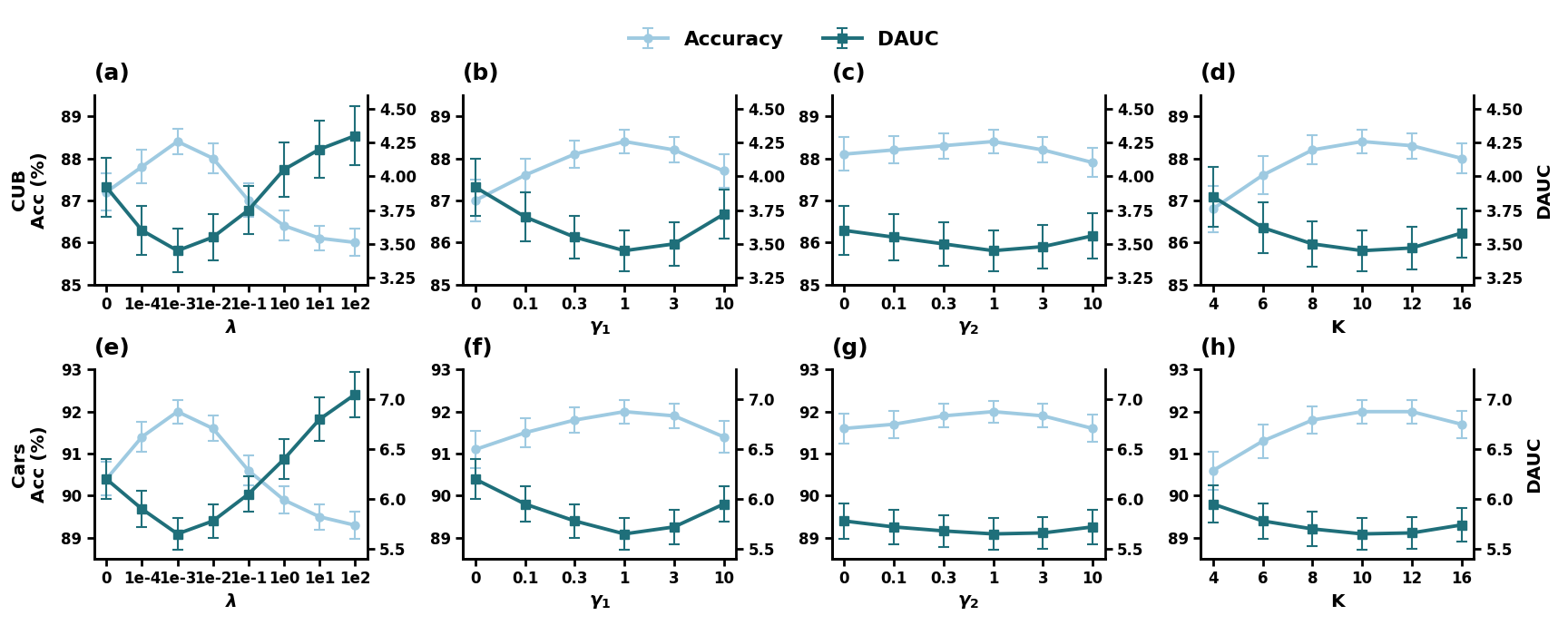}
    \caption{Hyperparameter sensitivity on CUB-200-2011 and Standford Cars.}
    \label{sens}
\end{figure}

\subsection{Quantitative Results}
Table~\ref{tab:main_cub_cars_sanity_colored} reports the predictive comparison on CUB-200-2011 and Stanford Cars where AMP consistently achieves the highest top-1 accuracy among intrinsically interpretable models across all backbones. On CUB-200-2011 with ResNet50, AMP reaches 88.4\% accuracy to surpass the strongest interpretable baseline MGProto at 86.6\% while remaining competitive with the black-box model PMG at 89.2\%. Stanford Cars exhibits a similar pattern with AMP achieving 92.0\% to exceed MGProto at 90.5\% and approach PMG at 93.6\%. Furthermore, AMP maintains stable state-of-the-art interpretable performance across VGG16, ResNet34, and DenseNet161 on both datasets. These results indicate that enforcing geometric diversity avoids an accuracy penalty and instead improves discrimination by preventing redundant prototype usage.

Table~\ref{tab:interpretability_metrics_sanity_colored} evaluates interpretability through Consistency, Stability, OIRR, and DAUC. AMP achieves the best scores across all four metrics on both datasets. On CUB-200-2011, AMP reaches 76.80 Consistency and 49.20 Stability to surpass the previous bests of MGProto and CBC, while attaining 28.10 OIRR and 3.45 DAUC to outperform ProtoArgNet and ProtoKNN. AMP similarly sets state-of-the-art interpretability records on Stanford Cars by achieving 50.20 Consistency, 76.40 Stability, 33.50 OIRR, and 5.65 DAUC. Overall, this quantitative evidence demonstrates that the Stiefel formulation effectively prevents prototype collapse and yields explanations that are stable across images and causally aligned with model decisions.

\subsection{Qualitative Results}
Fig~\ref{exp} visually demonstrates that AMP avoids prototype collapse and yields faithful compositional explanations on the CUB 200 2011 and Stanford Cars datasets. The model leverages geometric constraints to successfully discover diverse semantic parts such as the head and wing of a bird or the grille and wheel of a car. Visualizing each active Stiefel basis direction alongside its nearest training exemplar and spatial activation map confirms the efficacy of dynamic rank calibration. The exact contribution of each localized part is strictly quantified and cumulatively summed to derive the final class evidence score.

\subsection{Sensitivity Analysis}

We analyze the sensitivity of AMP to sparsity strength $\lambda$, spatial entropy weight $\gamma_1$, overlap suppression weight $\gamma_2$, and subspace size $K$ as illustrated in Fig~\ref{sens}. Across both datasets, A moderate $\lambda$ yields a U shaped trend in faithfulness metrics to optimally prune redundancy without collapsing discriminative capacity. Similarly, intermediate values for $\gamma_1$ and $\gamma_2$ resolve rotational ambiguity by encouraging localized and distinct evidence while avoiding the diffuse attention or artificial disjointness caused by extreme settings. Finally, an intermediate subspace size $K$ ensures sufficient representational capacity without the unconstrained redundancy of overly large bases. Overall, the broad optimal regions confirm that AMP is practically robust and consistently delivers causally faithful explanations without fragile tuning.

\begin{table*}[t]
\centering
\scriptsize
\caption{Ablation study on CUB-200-2011 and Stanford Cars. Best results are highlighted as \colorbox{first}{\textbf{first}}, \colorbox{second}{second} and \colorbox{third}{third}.}
\label{tab:ablation_study_sanity_colored}

\setlength{\tabcolsep}{3.5pt}
\renewcommand{\arraystretch}{1.08}

\begin{tabular}{l ccccc ccccc}
\toprule
\multirow{2}{*}{\centering\raisebox{-0.65ex}{\textbf{Method}}}
& \multicolumn{5}{c}{\textbf{CUB-200-2011}}
& \multicolumn{5}{c}{\textbf{Stanford Cars}} \\
\cmidrule(lr){2-6}\cmidrule(lr){7-11}
& \textbf{Acc} & \textbf{Cons.} & \textbf{Stab.} & \textbf{OIRR} & \textbf{DAUC}
& \textbf{Acc} & \textbf{Cons.} & \textbf{Stab.} & \textbf{OIRR} & \textbf{DAUC} \\
\midrule

w/o Stiefel
& 85.2 & 65.0 & 44.2 & 36.5 & 4.95
& 90.2 & 41.2 & 72.3 & 40.8 & 6.95 \\

w/o $\Sigma_c$
& \cellcolor{third}87.9 & \cellcolor{second}75.8 & 47.6 & \cellcolor{second}28.9 & \cellcolor{second}3.55
& \cellcolor{second}91.8 & \cellcolor{third}48.7 & \cellcolor{second}76.0 & 35.4 & \cellcolor{third}5.85 \\

w/o $\mathcal{L}_{SEM}$
& 87.0 & 73.2 & \cellcolor{third}47.8 & 30.4 & 3.92
& 91.1 & 46.2 & 74.5 & \cellcolor{third}35.0 & 6.20 \\

w/o $\mathcal{L}_{overlap}$
& \cellcolor{second}88.1 & \cellcolor{third}74.6 & \cellcolor{second}48.6 & \cellcolor{third}29.1 & \cellcolor{third}3.60
& \cellcolor{third}91.6 & \cellcolor{second}49.0 & \cellcolor{third}75.7 & \cellcolor{second}34.0 & \cellcolor{second}5.78 \\

\midrule
AMP (Full)
& \cellcolor{first}\textbf{88.4} & \cellcolor{first}\textbf{76.8} & \cellcolor{first}\textbf{49.2} & \cellcolor{first}\textbf{28.1} & \cellcolor{first}\textbf{3.45}
& \cellcolor{first}\textbf{92.0} & \cellcolor{first}\textbf{50.2} & \cellcolor{first}\textbf{76.4} & \cellcolor{first}\textbf{33.5} & \cellcolor{first}\textbf{5.65} \\
\bottomrule
\end{tabular}%
\end{table*}

\subsection{Ablation Study}
Table~\ref{tab:ablation_study_sanity_colored} ablates the key components of AMP on CUB-200-2011 and Stanford Cars. The results demonstrate that removing the Stiefel manifold constraint causes the most severe performance degradation by dropping CUB-200-2011 accuracy from 88.4 to 85.2, decreasing Consistency from 76.8 to 65.0, and raising DAUC to 4.95, confirming that hard orthogonality is essential to preserve representational capacity and prevent collapse. Similarly, omitting the dynamic capacity matrix $\Sigma_c$ slightly reduces accuracy and worsens interpretability by increasing OIRR on Stanford Cars from 33.5 to 35.4, underscoring the necessity of rank calibration for pruning redundant dimensions without sacrificing discriminative power. Furthermore, discarding the spatial gauge-fixing terms degrades both performance and faithfulness. Specifically, removing Spatial Entropy Minimization diffuses attention to drop CUB accuracy to 87.0 and raise DAUC to 3.92, while eliminating the Spatial Overlap Penalty fails to separate part evidence and negatively impacts OIRR. Ultimately, retaining all components enables AMP to achieve the best overall balance, delivering the highest accuracy and the most reliable explanations across both datasets.

\subsection{Human Evaluation}

\begin{figure}
    \centering
    \includegraphics[width=1\linewidth]{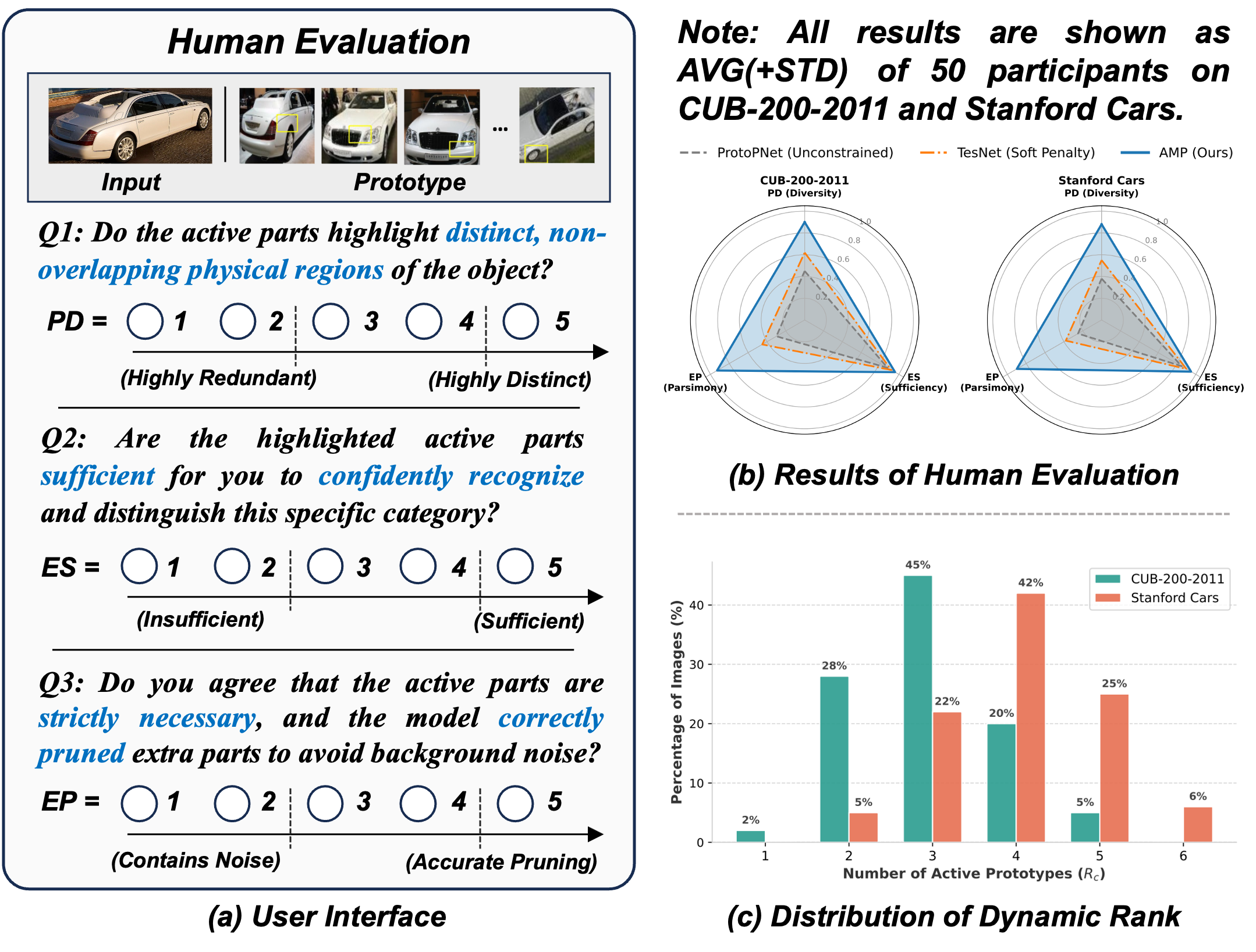}
    \caption{Human evaluation results and dynamic rank distribution.}
    \label{human}
\end{figure}

To assess the qualitative interpretability of AMP across three key dimensions and validate our dynamic rank design, we conducted a human evaluation study with 50 participants. As outlined in Fig~\ref{human}(a), we defined Part Diversity to measure region distinctness, Evidence Sufficiency to evaluate recognition confidence, and Explanation Parsimony to verify noise pruning. The results (Fig~\ref{human}(b)) demonstrate that AMP significantly outperforms both ProtoPNet and TesNet on the evaluated datasets, indicating that the Stiefel manifold effectively prevents redundancy while dynamic calibration ensures concise explanations. Furthermore, these human assessments align with the distribution of active prototypes (Fig~\ref{human}(c)). The data reveals that AMP adaptively adjusts its complexity according to the dataset: most CUB images activate three prototypes, whereas Stanford Cars typically require four. This adaptive behavior confirms that our dynamic rank mechanism successfully identifies the optimal number of semantic parts necessary for different categories, ultimately driving the high sufficiency and parsimony scores reported by the evaluators.

\section{Conclusion}
In this paper, we identified that the pervasive issue of prototype collapse in interpretable recognition is not merely an architectural flaw, but a geometric inevitability driven by the Neural Collapse under standard cross-entropy optimization. To resolve this, we introduce Adaptive Manifold Prototypes (AMP), which formulates class prototypes as orthonormal bases on the Stiefel manifold to mathematically preclude rank-one degeneracy. By incorporating proximal rank calibration and spatial gauge-fixing regularizers, AMP adaptively prunes redundancy while ensuring localized and semantically distinct part discovery. Extensive evaluations on fine-grained benchmarks demonstrate that AMP achieves state-of-the-art predictive performance while delivering exceptionally stable and causally faithful explanations. Ultimately, our work underscores a necessary paradigm shift for inherently interpretable AI: robust compositional reasoning demands strict geometric boundaries rather than heuristic soft penalties.



%
%
\bibliographystyle{splncs04}
\bibliography{main}

@String(ICML  = {Int. Conf. Mach. Learn.})

@String(AAAI  = {AAAI})

@String(ICML  = {ICML})

@article{rudin2019stop,
  title={Stop explaining black box machine learning models for high stakes decisions and use interpretable models instead},
  author={Rudin, Cynthia},
  journal={Nature Machine Intelligence},
  volume={1},
  number={5},
  pages={206--215},
  year={2019},
  publisher={Nature Publishing Group UK London}
}

@article{biederman1987recognition,
  title={Recognition-by-components: a theory of human image understanding.},
  author={Biederman, Irving},
  journal={Psychological Review},
  volume={94},
  number={2},
  pages={115},
  year={1987},
  publisher={American Psychological Association}
}

@article{chen2019looks,
  title={This looks like that: deep learning for interpretable image recognition},
  author={Chen, Chaofan and Li, Oscar and Tao, Daniel and Barnett, Alina and Rudin, Cynthia and Su, Jonathan K},
  journal={Advances in Neural Information Processing Systems},
  volume={32},
  year={2019}
}

@article{gautam2022protovae,
  title={Protovae: A trustworthy self-explainable prototypical variational model},
  author={Gautam, Srishti and Boubekki, Ahcene and Hansen, Stine and Salahuddin, Suaiba and Jenssen, Robert and H{\"o}hne, Marina and Kampffmeyer, Michael},
  journal={Advances in Neural Information Processing Systems},
  volume={35},
  pages={17940--17952},
  year={2022}
}

@inproceedings{parise2025looks,
  title={This Looks Like that and that and that: Multi-objective Optimization for Diverse Prototype Learning},
  author={Parise, Alec and Namee, Brian Mac},
  booktitle={International Conference on Innovative Techniques and Applications of Artificial Intelligence},
  pages={49--62},
  year={2025},
}

@inproceedings{zhang2014part,
  title={Part-based R-CNNs for fine-grained category detection},
  author={Zhang, Ning and Donahue, Jeff and Girshick, Ross and Darrell, Trevor},
  booktitle={European Conference on Computer Vision},
  pages={834--849},
  year={2014},
}

@inproceedings{branson2014bird,
	title = {Improved Bird Species Recognition Using Pose Normalized Deep Convolutional Nets},
	author = {Branson, Steve and Van Horn, Grant and Perona, Pietro and Belongie, Serge},
	year = {2014},
	booktitle = {Proceedings of the British Machine Vision Conference},
}

@article{markou2024guiding,
  title={Guiding neural collapse: Optimising towards the nearest simplex equiangular tight frame},
  author={Markou, Evan and Ajanthan, Thalaiyasingam and Gould, Stephen},
  journal={Advances in Neural Information Processing Systems},
  volume={37},
  pages={35544--35573},
  year={2024}
}

@article{papyan2020prevalence,
  title={Prevalence of neural collapse during the terminal phase of deep learning training},
  author={Papyan, Vardan and Han, XY and Donoho, David L},
  journal={Proceedings of the National Academy of Sciences},
  volume={117},
  number={40},
  pages={24652--24663},
  year={2020},
  publisher={National Academy of Sciences}
}

@article{edelman1998geometry,
  title={The geometry of algorithms with orthogonality constraints},
  author={Edelman, Alan and Arias, Tom{\'a}s A and Smith, Steven T},
  journal={SIAM Journal on Matrix Analysis and Applications},
  volume={20},
  number={2},
  pages={303--353},
  year={1998},
  publisher={SIAM}
}

@article{wen2013feasible,
  title={A feasible method for optimization with orthogonality constraints},
  author={Wen, Zaiwen and Yin, Wotao},
  journal={Mathematical Programming},
  volume={142},
  number={1},
  pages={397--434},
  year={2013},
}

@inproceedings{Li2020Efficient,
    title={Efficient Riemannian Optimization on the Stiefel Manifold via the Cayley Transform},
    author={Jun Li and Fuxin Li and Sinisa Todorovic},
    booktitle={International Conference on Learning Representations},
    year={2020},
}

@inproceedings{selvaraju2017grad,
  title={Grad-cam: Visual explanations from deep networks via gradient-based localization},
  author={Selvaraju, Ramprasaath R and Cogswell, Michael and Das, Abhishek and Vedantam, Ramakrishna and Parikh, Devi and Batra, Dhruv},
  booktitle={Proceedings of the IEEE International Conference on Computer Vision},
  pages={618--626},
  year={2017}
}

@article{adebayo2018sanity,
  title={Sanity checks for saliency maps},
  author={Adebayo, Julius and Gilmer, Justin and Muelly, Michael and Goodfellow, Ian and Hardt, Moritz and Kim, Been},
  journal={Advances in Neural Information Processing Systems},
  volume={31},
  year={2018}
}

@inproceedings{slack2020fooling,
  title={Fooling lime and shap: Adversarial attacks on post hoc explanation methods},
  author={Slack, Dylan and Hilgard, Sophie and Jia, Emily and Singh, Sameer and Lakkaraju, Himabindu},
  booktitle={Proceedings of the AAAI/ACM Conference on AI, Ethics, and Society},
  pages={180--186},
  year={2020}
}

@article{alvarez2018towards,
  title={Towards robust interpretability with self-explaining neural networks},
  author={Alvarez Melis, David and Jaakkola, Tommi},
  journal={Advances in Neural Information Processing Systems},
  volume={31},
  year={2018}
}

@inproceedings{wang2021interpretable,
  title={Interpretable image recognition by constructing transparent embedding space},
  author={Wang, Jiaqi and Liu, Huafeng and Wang, Xinyue and Jing, Liping},
  booktitle={Proceedings of the IEEE/CVF International Conference on Computer Vision},
  pages={895--904},
  year={2021}
}

@inproceedings{donnelly2022deformable,
  title={Deformable protopnet: An interpretable image classifier using deformable prototypes},
  author={Donnelly, Jon and Barnett, Alina Jade and Chen, Chaofan},
  booktitle={Proceedings of the IEEE/CVF Conference on Computer Vision and Pattern Recognition},
  pages={10265--10275},
  year={2022}
}

@inproceedings{hoffmann2021looks,
    title={This Looks Like That... Does it? Shortcomings of Latent Space Prototype Interpretability in Deep Networks}, 
    author={Adrian Hoffmann and Claudio Fanconi and Rahul Rade and Jonas Kohler},
    year={2021},
    booktitle={ICML 2021 Workshop on Theoretic Foundation, Criticism, and Application Trend of Explainable AI},
}

@article{zhu2021geometric,
  title={A geometric analysis of neural collapse with unconstrained features},
  author={Zhu, Zhihui and Ding, Tianyu and Zhou, Jinxin and Li, Xiao and You, Chong and Sulam, Jeremias and Qu, Qing},
  journal={Advances in Neural Information Processing Systems},
  volume={34},
  pages={29820--29834},
  year={2021}
}

@incollection{kindermans2019reliability,
  title={The (un) reliability of saliency methods},
  author={Kindermans, Pieter-Jan and Hooker, Sara and Adebayo, Julius and Alber, Maximilian and Sch{\"u}tt, Kristof T and D{\"a}hne, Sven and Erhan, Dumitru and Kim, Been},
  booktitle={Explainable AI: Interpreting, Explaining and Visualizing Deep Learning},
  pages={267--280},
  year={2019},
}

@article{hooker2019benchmark,
  title={A benchmark for interpretability methods in deep neural networks},
  author={Hooker, Sara and Erhan, Dumitru and Kindermans, Pieter-Jan and Kim, Been},
  journal={Advances in Neural Information Processing Systems},
  volume={32},
  year={2019}
}

@article{yeh2019fidelity,
  title={On the (in) fidelity and sensitivity of explanations},
  author={Yeh, Chih-Kuan and Hsieh, Cheng-Yu and Suggala, Arun and Inouye, David I and Ravikumar, Pradeep K},
  journal={Advances in Neural Information Processing Systems},
  volume={32},
  year={2019}
}

@inproceedings{ghorbani2019interpretation,
  title={Interpretation of neural networks is fragile},
  author={Ghorbani, Amirata and Abid, Abubakar and Zou, James},
  booktitle={Proceedings of the AAAI Conference on Artificial Intelligence},
  volume={33},
  number={01},
  pages={3681--3688},
  year={2019}
}

@article{dombrowski2019explanations,
  title={Explanations can be manipulated and geometry is to blame},
  author={Dombrowski, Ann-Kathrin and Alber, Maximillian and Anders, Christopher and Ackermann, Marcel and M{\"u}ller, Klaus-Robert and Kessel, Pan},
  journal={Advances in Neural Information Processing Systems},
  volume={32},
  year={2019}
}

@inproceedings{aivodji2019fairwashing,
  title={Fairwashing: the risk of rationalization},
  author={A{\"\i}vodji, Ulrich and Arai, Hiromi and Fortineau, Olivier and Gambs, S{\'e}bastien and Hara, Satoshi and Tapp, Alain},
  booktitle={International Conference on Machine Learning},
  pages={161--170},
  year={2019},
  organization={PMLR}
}

@article{ilyas2019adversarial,
  title={Adversarial examples are not bugs, they are features},
  author={Ilyas, Andrew and Santurkar, Shibani and Tsipras, Dimitris and Engstrom, Logan and Tran, Brandon and Madry, Aleksander},
  journal={Advances in Neural Information Processing Systems},
  volume={32},
  year={2019}
}

@inproceedings{locatello2019challenging,
  title={Challenging common assumptions in the unsupervised learning of disentangled representations},
  author={Locatello, Francesco and Bauer, Stefan and Lucic, Mario and Raetsch, Gunnar and Gelly, Sylvain and Sch{\"o}lkopf, Bernhard and Bachem, Olivier},
  booktitle={International Conference on Machine Learning},
  pages={4114--4124},
  year={2019},
  organization={PMLR}
}

@inproceedings{ribeiro2016should,
  title={" Why should i trust you?" Explaining the predictions of any classifier},
  author={Ribeiro, Marco Tulio and Singh, Sameer and Guestrin, Carlos},
  booktitle={Proceedings of the 22nd ACM SIGKDD International Conference on Knowledge Discovery and Data Mining},
  pages={1135--1144},
  year={2016}
}

@article{lundberg2017unified,
  title={A unified approach to interpreting model predictions},
  author={Lundberg, Scott M and Lee, Su-In},
  journal={Advances in Neural Information Processing Systems},
  volume={30},
  year={2017}
}

@article{geirhos2020shortcut,
  title={Shortcut learning in deep neural networks},
  author={Geirhos, Robert and Jacobsen, J{\"o}rn-Henrik and Michaelis, Claudio and Zemel, Richard and Brendel, Wieland and Bethge, Matthias and Wichmann, Felix A},
  journal={Nature Machine Intelligence},
  volume={2},
  number={11},
  pages={665--673},
  year={2020},
  publisher={Nature Publishing Group UK London}
}

@techreport{wah2011caltech,
  title={The caltech-ucsd birds-200-2011 dataset},
  author={Wah, Catherine and Branson, Steve and Welinder, Peter and Perona, Pietro and Belongie, Serge and others},
  year={2011},
  institution={Technical Report CNS-TR-2011-001, California Institute of Technology}
}

@inproceedings{krause20133d,
  title={3d object representations for fine-grained categorization},
  author={Krause, Jonathan and Stark, Michael and Deng, Jia and Fei-Fei, Li},
  booktitle={Proceedings of the IEEE International Conference on Computer Vision Workshops},
  pages={554--561},
  year={2013}
}

@inproceedings{simonyan2015very,
  title={Very deep convolutional networks for large-scale image recognition},
  author={Simonyan, K and Zisserman, A},
  booktitle={International Conference on Learning Representations},
  year={2015}
}

@inproceedings{he2016deep,
  title={Deep residual learning for image recognition},
  author={He, Kaiming and Zhang, Xiangyu and Ren, Shaoqing and Sun, Jian},
  booktitle={Proceedings of the IEEE Conference on Computer Vision and Pattern Recognition},
  pages={770--778},
  year={2016}
}

@inproceedings{huang2017densely,
  title={Densely connected convolutional networks},
  author={Huang, Gao and Liu, Zhuang and Van Der Maaten, Laurens and Weinberger, Kilian Q},
  booktitle={Proceedings of the IEEE Conference on Computer Vision and Pattern Recognition},
  pages={4700--4708},
  year={2017}
}

@inproceedings{deng2009imagenet,
  title={Imagenet: A large-scale hierarchical image database},
  author={Deng, Jia and Dong, Wei and Socher, Richard and Li, Li-Jia and Li, Kai and Fei-Fei, Li},
  booktitle={IEEE Conference on Computer Vision and Pattern Recognition},
  pages={248--255},
  year={2009},
}

@inproceedings{van2018inaturalist,
  title={The inaturalist species classification and detection dataset},
  author={Van Horn, Grant and Mac Aodha, Oisin and Song, Yang and Cui, Yin and Sun, Chen and Shepard, Alex and Adam, Hartwig and Perona, Pietro and Belongie, Serge},
  booktitle={Proceedings of the IEEE Conference on Computer Vision and Pattern Recognition},
  pages={8769--8778},
  year={2018}
}

@inproceedings{huang2023evaluation,
  title={Evaluation and improvement of interpretability for self-explainable part-prototype networks},
  author={Huang, Qihan and Xue, Mengqi and Huang, Wenqi and Zhang, Haofei and Song, Jie and Jing, Yongcheng and Song, Mingli},
  booktitle={Proceedings of the IEEE/CVF International Conference on Computer Vision},
  pages={2011--2020},
  year={2023}
}

@inproceedings{fu2017look,
  title={Look closer to see better: Recurrent attention convolutional neural network for fine-grained image recognition},
  author={Fu, Jianlong and Zheng, Heliang and Mei, Tao},
  booktitle={Proceedings of the IEEE Conference on Computer Vision and Pattern Recognition},
  pages={4438--4446},
  year={2017}
}

@inproceedings{yang2018learning,
  title={Learning to navigate for fine-grained classification},
  author={Yang, Ze and Luo, Tiange and Wang, Dong and Hu, Zhiqiang and Gao, Jun and Wang, Liwei},
  booktitle={Proceedings of the European Conference on Computer Vision},
  pages={420--435},
  year={2018}
}

@inproceedings{du2020fine,
  title={Fine-grained visual classification via progressive multi-granularity training of jigsaw patches},
  author={Du, Ruoyi and Chang, Dongliang and Bhunia, Ayan Kumar and Xie, Jiyang and Ma, Zhanyu and Song, Yi-Zhe and Guo, Jun},
  booktitle={European Conference on Computer Vision},
  pages={153--168},
  year={2020},
}

@inproceedings{ukai2023this,
    title={This Looks Like It Rather Than That: Proto{KNN} For Similarity-Based Classifiers},
    author={Yuki Ukai and Tsubasa Hirakawa and Takayoshi Yamashita and Hironobu Fujiyoshi},
    booktitle={International Conference on Learning Representations},
    year={2023}
}

@article{wang2025mixture,
  title={Mixture of gaussian-distributed prototypes with generative modelling for interpretable and trustworthy image recognition},
  author={Wang, Chong and Chen, Yuanhong and Liu, Fengbei and Liu, Yuyuan and McCarthy, Davis James and Frazer, Helen and Carneiro, Gustavo},
  journal={IEEE Transactions on Pattern Analysis and Machine Intelligence},
  year={2025},
  publisher={IEEE}
}

@inproceedings{saralajew2025robust,
  title={A robust prototype-based network with interpretable RBF classifier foundations},
  author={Saralajew, Sascha and Rana, Ashish and Villmann, Thomas and Shaker, Ammar},
  booktitle={Proceedings of the AAAI Conference on Artificial Intelligence},
  volume={39},
  number={19},
  pages={20273--20282},
  year={2025}
}

@inproceedings{ayoobi2025protoargnet,
  title={ProtoArgNet: Interpretable image classification with super-prototypes and argumentation},
  author={Ayoobi, Hamed and Potyka, Nico and Toni, Francesca},
  booktitle={Proceedings of the AAAI Conference on Artificial Intelligence},
  volume={39},
  number={2},
  pages={1791--1799},
  year={2025}
}

@book{absil2008optimization,
  title={Optimization algorithms on matrix manifolds},
  author={Absil, P-A and Mahony, Robert and Sepulchre, Rodolphe},
  year={2008},
  publisher={Princeton University Press}
}

@article{bonnabel2013stochastic,
  title={Stochastic gradient descent on Riemannian manifolds},
  author={Bonnabel, Silvere},
  journal={IEEE Transactions on Automatic Control},
  volume={58},
  number={9},
  pages={2217--2229},
  year={2013},
  publisher={IEEE}
}

@article{zhang2016riemannian,
  title={Riemannian SVRG: Fast stochastic optimization on Riemannian manifolds},
  author={Zhang, Hongyi and J Reddi, Sashank and Sra, Suvrit},
  journal={Advances in Neural Information Processing Systems},
  volume={29},
  year={2016}
}

@inproceedings{huang2018orthogonal,
  title={Orthogonal weight normalization: Solution to optimization over multiple dependent stiefel manifolds in deep neural networks},
  author={Huang, Lei and Liu, Xianglong and Lang, Bo and Yu, Adams and Wang, Yongliang and Li, Bo},
  booktitle={Proceedings of the AAAI Conference on Artificial Intelligence},
  volume={32},
  number={1},
  year={2018}
}

@article{bansal2018can,
  title={Can we gain more from orthogonality regularizations in training deep networks?},
  author={Bansal, Nitin and Chen, Xiaohan and Wang, Zhangyang},
  journal={Advances in Neural Information Processing Systems},
  volume={31},
  year={2018}
}

@inproceedings{roy2018geometry,
  title={Geometry aware constrained optimization techniques for deep learning},
  author={Roy, Soumava Kumar and Mhammedi, Zakaria and Harandi, Mehrtash},
  booktitle={Proceedings of the IEEE Conference on Computer Vision and Pattern Recognition},
  pages={4460--4469},
  year={2018}
}

@article{pandey2022disentangled,
  title={Disentangled representation learning and generation with manifold optimization},
  author={Pandey, Arun and Fanuel, Micha{\"e}l and Schreurs, Joachim and Suykens, Johan AK},
  journal={Neural Computation},
  volume={34},
  number={10},
  pages={2009--2036},
  year={2022}
}

@inproceedings{cogswell2016decov,
    title={Reducing overfitting in deep networks by decorrelating representations},
    author={Cogswell, Michael and Ahmed, Faruk and Girshick, Ross and Zitnick, Larry and Batra, Dhruv},
    booktitle={International Conference on Learning Representations},
    year={2016}
}

@inproceedings{nauta2021neural,
  title={Neural prototype trees for interpretable fine-grained image recognition},
  author={Nauta, Meike and Van Bree, Ron and Seifert, Christin},
  booktitle={Proceedings of the IEEE/CVF Conference on Computer Vision and Pattern Recognition},
  pages={14933--14943},
  year={2021}
}

@article{jia2025geodesic,
  title={Geodesic prototype matching via diffusion maps for interpretable fine-grained recognition},
  author={Jia, Junhao and Liu, Yunyou and Sun, Yifei and Chen, Huangwei and Qin, Feiwei and Wang, Changmiao and Peng, Yong},
  journal={arXiv preprint arXiv:2509.17050},
  year={2025}
}

@article{shalev2011stochastic,
  title={Stochastic Methods for l 1-regularized Loss Minimization},
  author={Shalev-Shwartz, Shai and Tewari, Ambuj},
  journal={Journal of Machine Learning Research},
  volume={12},
  pages={1865--1892},
  year={2011},
  publisher={JMLR. org}
}

@article{langford2008sparse,
  title={Sparse online learning via truncated gradient},
  author={Langford, John and Li, Lihong and Zhang, Tong},
  journal={Advances in Neural Information Processing Systems},
  volume={21},
  year={2008}
}

@article{ye2022optimization,
  title={Optimization on flag manifolds},
  author={Ye, Ke and Wong, Ken Sze-Wai and Lim, Lek-Heng},
  journal={Mathematical Programming},
  volume={194},
  number={1},
  pages={621--660},
  year={2022},
}

@article{parikh2014proximal,
  title={Proximal algorithms},
  author={Parikh, Neal and Boyd, Stephen},
  journal={Foundations and Trends in Optimization},
  volume={1},
  number={3},
  pages={127--239},
  year={2014}
}

@article{yaras2022neural,
  title={Neural collapse with normalized features: A geometric analysis over the riemannian manifold},
  author={Yaras, Can and Wang, Peng and Zhu, Zhihui and Balzano, Laura and Qu, Qing},
  journal={Advances in Neural Information Processing Systems},
  volume={35},
  pages={11547--11560},
  year={2022}
}

@article{jia2026unsupervised,
  title={Unsupervised Causal Prototypical Networks for De-biased Interpretable Dermoscopy Diagnosis},
  author={Jia, Junhao and Wu, Yueyi and Chen, Huangwei and Jing, Haodong and Wang, Haishuai and Bu, Jiajun and Wu, Lei},
  journal={arXiv preprint arXiv:2602.23752},
  year={2026}
}

@inproceedings{wei-zhu-2025-protolens,
    title = "{P}roto{L}ens: Advancing Prototype Learning for Fine-Grained Interpretability in Text Classification",
    author = "Wei, Bowen  and Zhu, Ziwei",
    booktitle = "Proceedings of the 63rd Annual Meeting of the Association for Computational Linguistics (Volume 1: Long Papers)",
    year = "2025",
    pages = "4503--4523",
}

@inproceedings{peng2026protots,
    title={Proto{TS}: Learning Hierarchical Prototypes for Explainable Time Series Forecasting},
    author={Ziheng Peng and Shijie Ren and Xinyue Gu and Linxiao Yang and Xiting Wang and Liang Sun},
    booktitle={International Conference on Learning Representations},
    year={2026}
}

@article{wang2025cross,
  title={Cross-and intra-image prototypical learning for multi-label disease diagnosis and interpretation},
  author={Wang, Chong and Liu, Fengbei and Chen, Yuanhong and Frazer, Helen and Carneiro, Gustavo},
  journal={IEEE Transactions on Medical Imaging},
  year={2025},
  publisher={IEEE}
}

@article{hong2023protorynet,
  title={Protorynet-interpretable text classification via prototype trajectories},
  author={Hong, Dat and Wang, Tong and Baek, Stephen},
  journal={Journal of Machine Learning Research},
  volume={24},
  number={264},
  pages={1--39},
  year={2023}
}

@inproceedings{ni2021interpreting,
  title={Interpreting convolutional sequence model by learning local prototypes with adaptation regularization},
  author={Ni, Jingchao and Chen, Zhengzhang and Cheng, Wei and Zong, Bo and Song, Dongjin and Liu, Yanchi and Zhang, Xuchao and Chen, Haifeng},
  booktitle={Proceedings of the 30th ACM International Conference on Information \& Knowledge Management},
  pages={1366--1375},
  year={2021}
}
\end{document}